\documentclass{article} 
\usepackage{nips13submit_e,times}
\usepackage{hyperref}
\usepackage{url}
\usepackage{graphicx}
\usepackage{subfigure}

\title{Face R-CNN}

\author{
Hao Wang\quad Zhifeng Li\thanks{Corresponding author}\quad Xing Ji\quad Yitong Wang \\
Tencent AI Lab, China\\
\texttt{\{hawelwang,michaelzfli,denisji,yitongwang\}@tencent.com} \\
}

\nipsfinalcopy 

\begin{document}

\maketitle

\begin{abstract}
Faster R-CNN is one of the most representative and successful methods for object detection, and has been becoming increasingly popular in various objection detection applications. In this report, we propose a robust deep face detection approach based on Faster R-CNN. In our approach, we exploit several new techniques including new multi-task loss function design, online hard example mining, and multi-scale training strategy to improve Faster R-CNN in multiple aspects. The proposed approach is well suited for face detection, so we call it \emph{Face R-CNN}. Extensive experiments are conducted on two most popular and challenging face detection benchmarks, FDDB and WIDER FACE, to demonstrate the superiority of the proposed approach over state-of-the-arts.
\end{abstract}

\section{Introduction}

Face detection is a fundamental problem in a large number of face-related applications in computer vision and pattern recognition. However, it still remains challenging due to significant variations of real-world face images. Figure 1 is a typical example, where the face images exhibit a large degree of variations caused by occlusion, scale, illumination, pose, and expression.

Face detection can be considered as a special case of objection detection. Recently, region-based convolutional neural networks \cite{RCNNs,RCNN,FastRCNN,FasterRCNN} have achieved state-of-the-art performance on generic object detection. Among the top-performing region-based CNN methods, Fast R-CNN \cite{FastRCNN} and Faster R-CNN \cite{FasterRCNN} are among the most popular ones. Fast R-CNN trains object detectors based on the regions of interest (RoIs), which are generated by region proposal methods such as selective search \cite{Selectivesearch}. Faster R-CNN inherits the basic framework of Fast R-CNN but offers an improvement in generating object proposals by constructing a Region Proposal Network (RPN). Faster R-CNN has very good advantages in effectiveness and efficiency, and thus acts as the leading framework in various detection tasks \cite{maskrcnn,FPN,OHEM}. Recently, a number of Faster R-CNN based methods \cite{facefrcnn,xm,deepir,cmsrcnn} have been developed for face detection task and demonstrated the state-of-the-art performance.

Though great progress has been achieved by the Faster R-CNN based methods \cite{facefrcnn,xm,deepir,cmsrcnn}, there are still some issues with these methods. A common problem with these methods is that they use the softmax loss function to supervise the learning of the deep features for face/non-face classification in the Fast R-CNN module (following the standard practice in Faster R-CNN). However, it has been shown that the softmax loss only encourages the inter-class separability of the learned features, but fails to encourage the intra-class compactness of the learned features. Previous studies \cite{deepid2,latent,centerloss,facenet} have shown that both the inter-class separability and intra-class compactness are very important for the discriminative power of the learned CNN features. In order to reduce the intra-class variations while enlarging the inter-class variations of the learned CNN features, we improve the Faster R-CNN framework by adding a newly developed loss function called center loss \cite{centerloss} to the original multi-task loss function. By adding the center loss, the intra-class variations of the learned features can be effectively reduced, and thus the discriminative power of the learned features can be enhanced accordingly. In order to further improve the detection accuracy, we also employ the online hard example mining (OHEM) technique \cite{OHEM} and the multi-scale training strategy in this study.

\begin{figure}
  \centering
  \includegraphics[width=10cm, keepaspectratio]{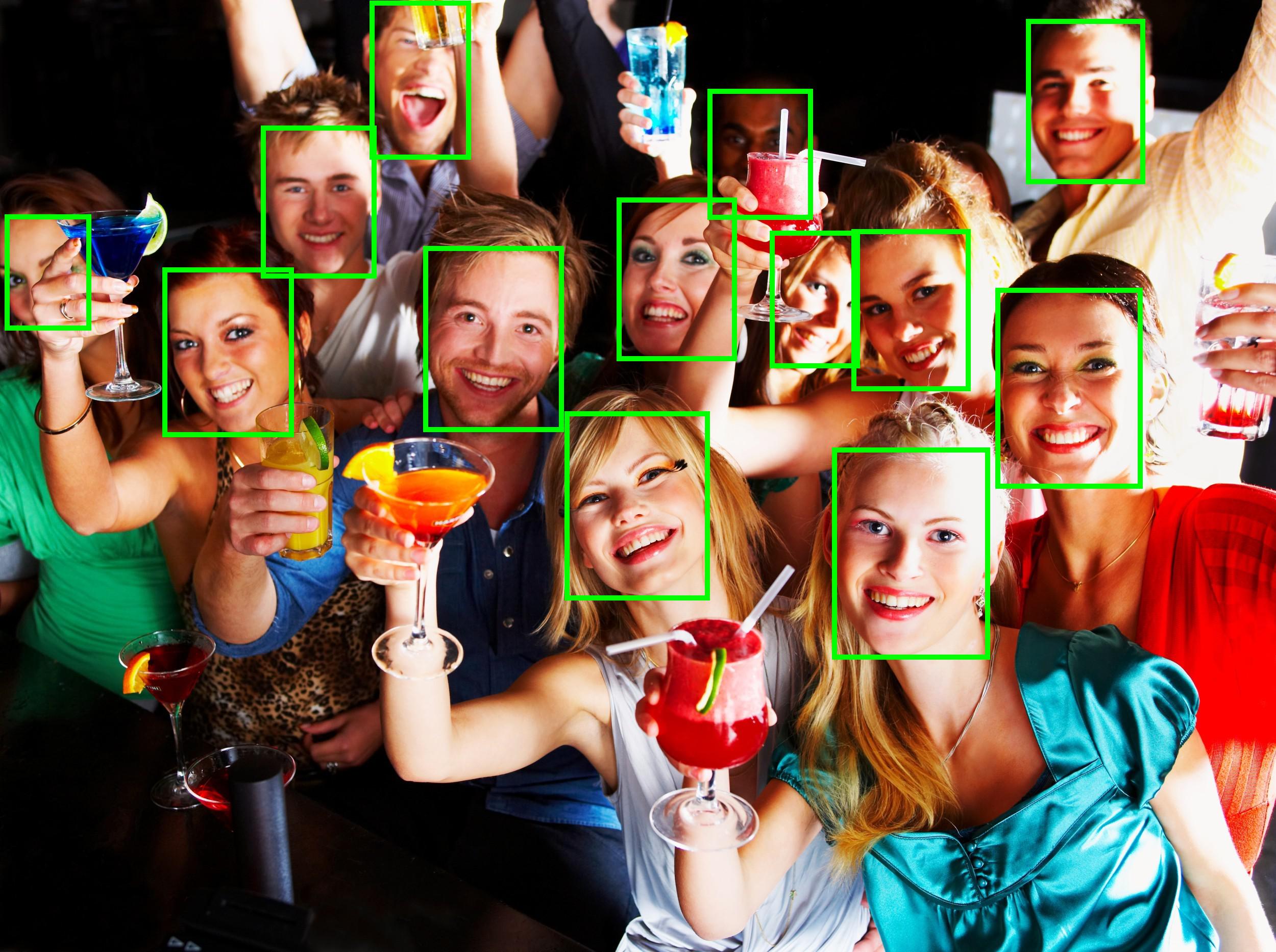}
  \caption{We show an example image for face detection, which has representative variability in occlusion, scale, illumination, pose,
  and expression. Detection results are displayed using the proposed face detector.}\label{1}
\end{figure}

The major contributions of this work are summarized as follows:

(1) Considering the specific property of face detection, we propose a Faster R-CNN based approach called \emph{Face R-CNN} for face detection by integrating several newly developed techniques including center loss, online hard example mining, and multi-scale training.

(2) The proposed approach differs from the available Faster R-CNN based face detection methods. First, this is the first attempt to use the center loss to reduce the large intra-class variations in face detection. Second, the use of online hard example mining in our approach differs from the others. By appropriately setting the ratio between positive hard samples and negative hard samples, the combination of OHEM and center loss can lead to better performance.

(3) The proposed approach consistently obtains superior performance over the state-of-the-arts on two public-domain face detection benchmarks (WIDER FACE dataset \cite{wider} and FDDB dataset \cite{fddb}).

\section{Related Work}

Face detection has been extensively studied in the past two decades
due to its increasing applications.
Early studies include cascade based methods
\cite{vj,vj1,vj2,vj3} and deformable part models (DPM) based methods \cite{dpmface,dpmface2,dpmface3}.
In 2004, Viola and Jones \cite{vj} proposed to learn the cascaded AdaBoost classifier based on Haar-like features for face detection.
After that, several works \cite{vj1,vj2,vj3} improve \cite{vj} by using more advanced
features and classifiers.
Besides the cascade based face detection methods, \cite{dpmface2,dpmface3,dpmface} propose to use deformable
part models (DPM). The idea of DPM is to define a face
with a collection of deformable parts and train a classifier to find these parts and their relationship. Recent advances in face detection mainly focus on the deep learning based approaches.
\cite{cascadeCNN,cascadeCNN2,spl} construct cascaded CNNs to learn face detectors with a coarse-to-fine
strategy. UnitBox \cite{unitbox} introduces the intersection-over-union (IoU) loss function, to directly minimize the
IoUs of the predictions and the ground-truths. More recently, several methods \cite{facefrcnn,xm,deepir,cmsrcnn} use the Faster R-CNN framework to improve the face detection performance. Most recently, \cite{HR} significantly improves the face detection performance over prior methods.

The methods of hard example mining have been demonstrated effective on object detection tasks. In \cite{OHEM}, the authors proposed an online hard example mining (OHEM) algorithm (trained with the Fast R-CNN framework) to improve the object detection performance.
\cite{xm,deepir,spl} also use hard example mining algorithms to boost the performance of face detection.

Loss function also plays a key role in deep CNNs. The most widely used loss function in deep CNNs is softmax loss function. However, the softmax loss function focuses on inter-class separability while neglects intra-class compactness in feature learning, which would affect the classification accuracy. To enhance discriminative ability of the learned features, contrastive loss \cite{deepid2,latent} and triplet loss \cite{facenet} have been recently used in deep CNNs. More recently, center loss \cite{centerloss} is proposed
by minimizing the distances between samples and the learned class centers. The center loss is
efficient, easy to implement, and can be well generalized to other visual tasks. In this report, we will use the center loss to improve the performance of
face detection.

\section{Face R-CNN}

Motivated by the great success of the state-of-the-art Faster R-CNN framework in detection tasks, we use it as a general platform to design our new model for face detection.
Considering the specific property of the face detection task, we improve the Faster R-CNN framework in multiple aspects. First, based on a newly developed loss function called center loss \cite{centerloss}, we design a new multi-task loss function in the Fast R-CNN model to supervise the learning of discriminative deep features for face/non-face classification. Second, in the learning of deep features, we use online hard example mining algorithm \cite{OHEM} to generate hard samples (in which the ratio of positive samples to negative samples is set to be 1:1) for subsequent processing. Third, we use multi-scale training strategy to help improve the detection performance. By integrating the above techniques, the proposed approach is well suited for face detection task, as supported by our experimental results. Hence we call it \emph{Face R-CNN}. The detailed algorithm of Face R-CNN is elaborated as follows.

\subsection{Faster R-CNN Based Architecture}

\begin{figure*}
  \centering
  \includegraphics[width=14cm, keepaspectratio]{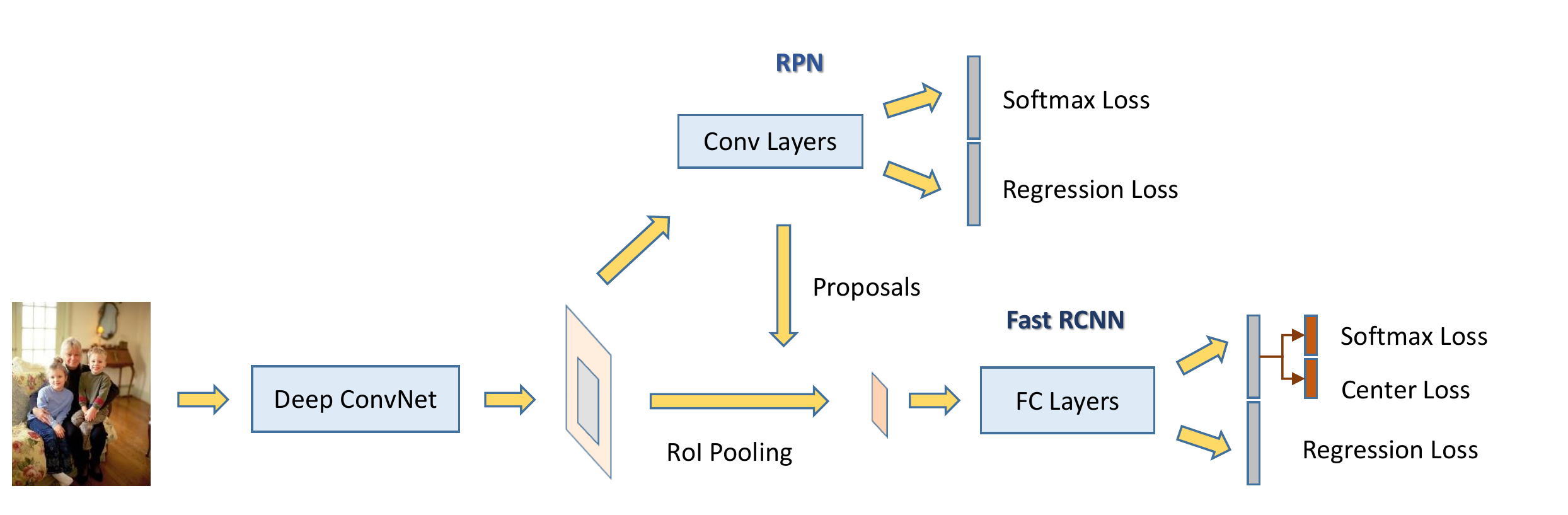}
  \caption{An overview of our Faster R-CNN based framework. We first extract
    conv features with a pre-trained ConvNet. Then, RPN produces candidate
    proposals and Fast R-CNN refines the results.
    We add the center loss as an auxiliary signal on the classification layer of Fast R-CNN.
    The underlying conv features are shared for PRN and Fast R-CNN.
    }\label{1}
\end{figure*}

We use the Faster R-CNN framework \cite{FasterRCNN} as the baseline architecture.
Figure 2 illustrates our architecture, which involves a basic ConvNet,
a Region Proposal Network (RPN) module, and a Fast R-CNN module.

The basic ConvNet processes an input image with a stack of convolutional layers and max pooling layers, and produces
the convolutional feature map. The RPN module produces a set of
rectangle region proposals that are very likely to contain the faces,
with a fully convolutional network built on top of the convolutional feature maps.
The loss layers of RPN contain a binary classification layer and a bounding box regression layer.
The generated region proposals are then fed into the Fast R-CNN module
and taken as the regions of interest (RoIs). The RoI pooling layer
processes the RoIs to extract the fixed-length feature vectors.
The final output are two sibling fully connected layers for classification and regression.

In the RPN stage, the learning of box-classification and box-regression applies an anchor-based method.
That is, a set of pre-defined boxes called anchors are used as box-references for regression.
The anchors are associate with multiple scales and aspect ratios, so that the bounding boxes of various shapes
can be covered.

The major difference between our framework and the typical Faster R-CNN lies in the Fast R-CNN module. Unlike \cite{FasterRCNN}, we design a new multi-task loss function based on a newly developed center loss to supervise the feature learning, as elaborated in the next section.

\subsection{Loss Function}

\subsubsection{Center Loss}

We first briefly review Center Loss \cite{centerloss}, which is a newly developed loss function and has demonstrated impressive results in face recognition task. The basic idea of center loss is to encourage the network to learn the discriminative features that minimize the intra-class variations while enlarging the inter-class variations.

The formulation of center loss is

\begin{equation}\label{1}
 L_c(x) = \frac{1}{2}\sum_{i=1}^{m}{\Vert{x_i}-{c_{y_i}}\Vert}_2^2,
\end{equation}

\noindent
where $x$ denotes the input feature vector, and ${c_{y_i}}$ denotes the ${y_i}$th class center. The centers are updated based on mini-batch in each iteration, so it can be easily trained
with standard SGD. For face detection task, there are only two centers representing faces and non-faces, respectively. Our purpose is to
minimize the intra-class variations.

One thing to note is that the center loss is supposed to be optimized jointly with
the softmax loss \cite{centerloss}. It has been shown that the center loss is very effective in reducing the intra-class variations as much as possible, and the softmax loss has some merits in maximizing the inter-class variations of the learned features. Therefore it is very reasonable to use a combination of center loss and softmax loss to pursue the discriminative features.

\begin{figure*}
  \centering

  \subfigure[Val: easy]
  {
    \label{fig:2:a}
    \includegraphics[width=4cm, keepaspectratio]{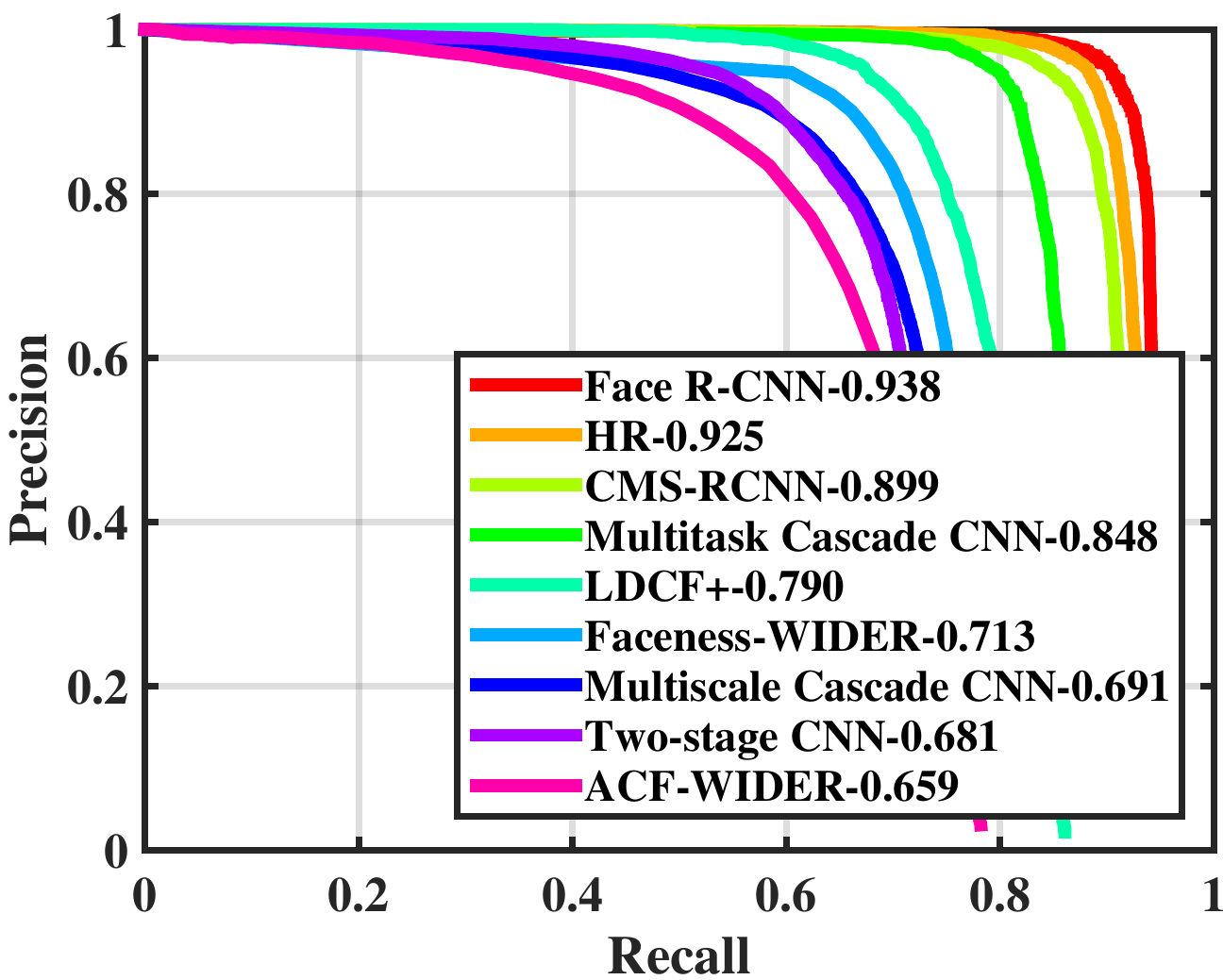}
  }
  \subfigure[Val: medium]
  {
  \label{fig:2:a}
  \includegraphics[width=4cm, keepaspectratio]{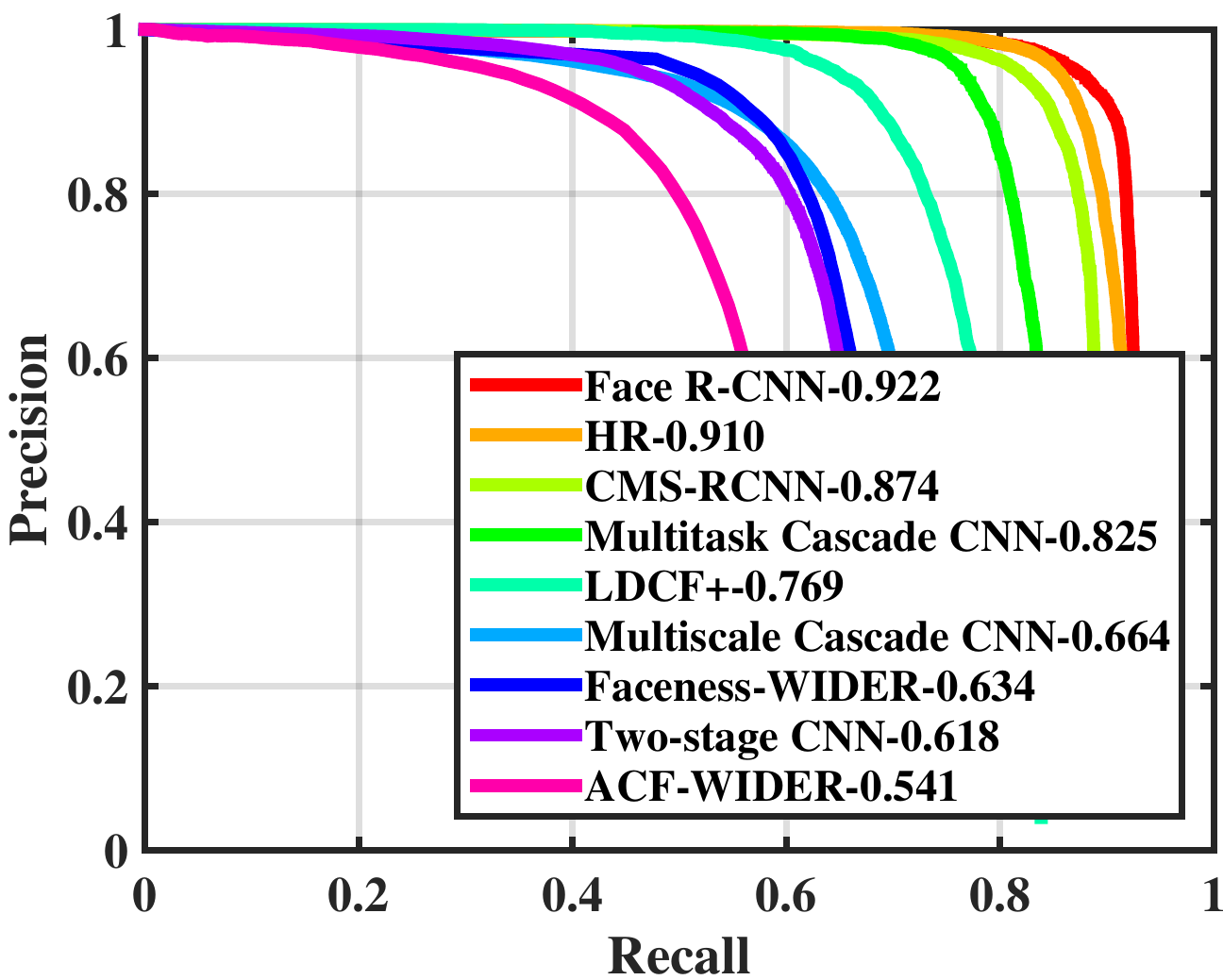}
  }
  \subfigure[Val: hard]
  {
  \label{fig:2:a}
  \includegraphics[width=4cm, keepaspectratio]{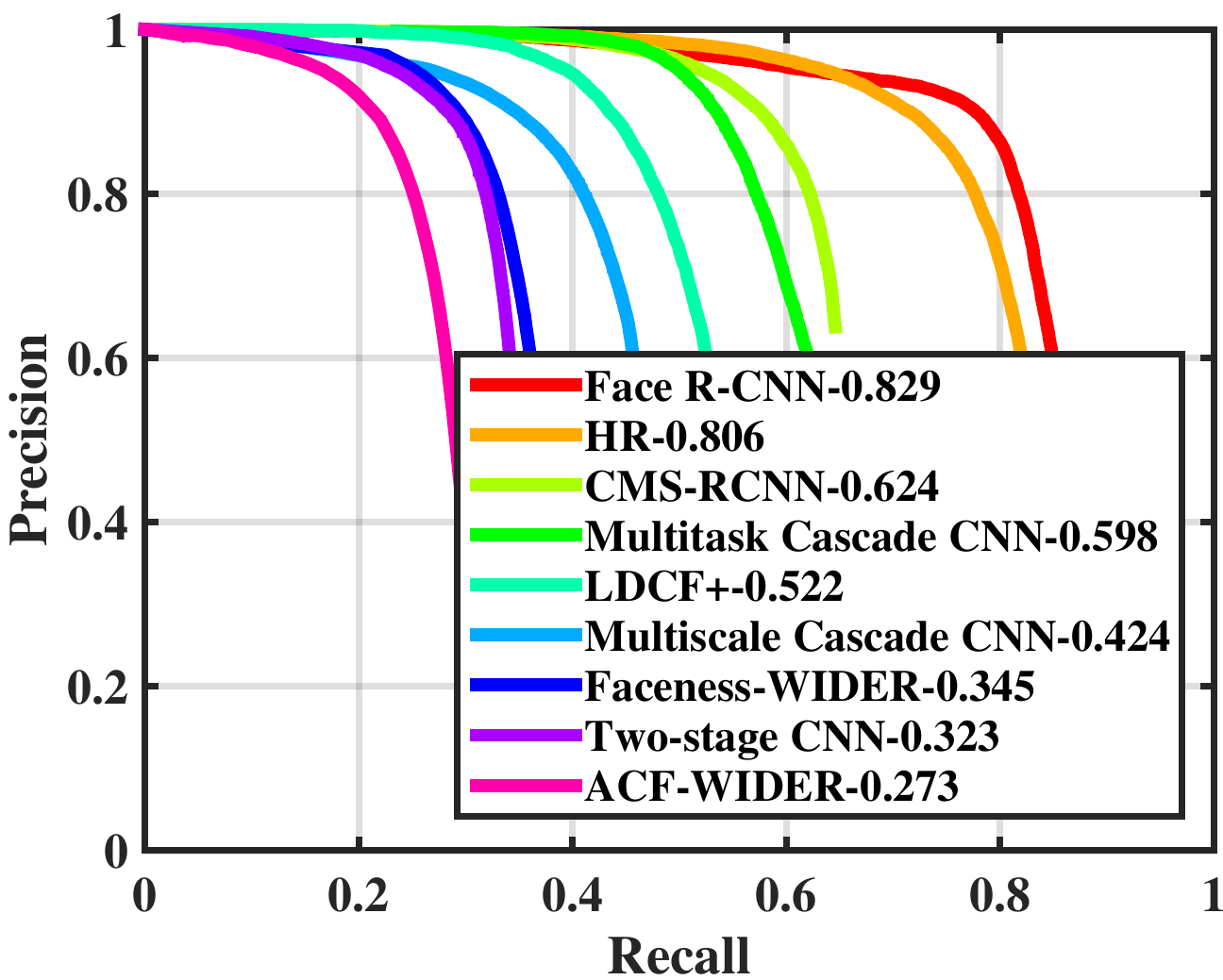}
  }

  \subfigure[Test: easy]
  {
    \label{fig:2:a}
    \includegraphics[width=4cm, keepaspectratio]{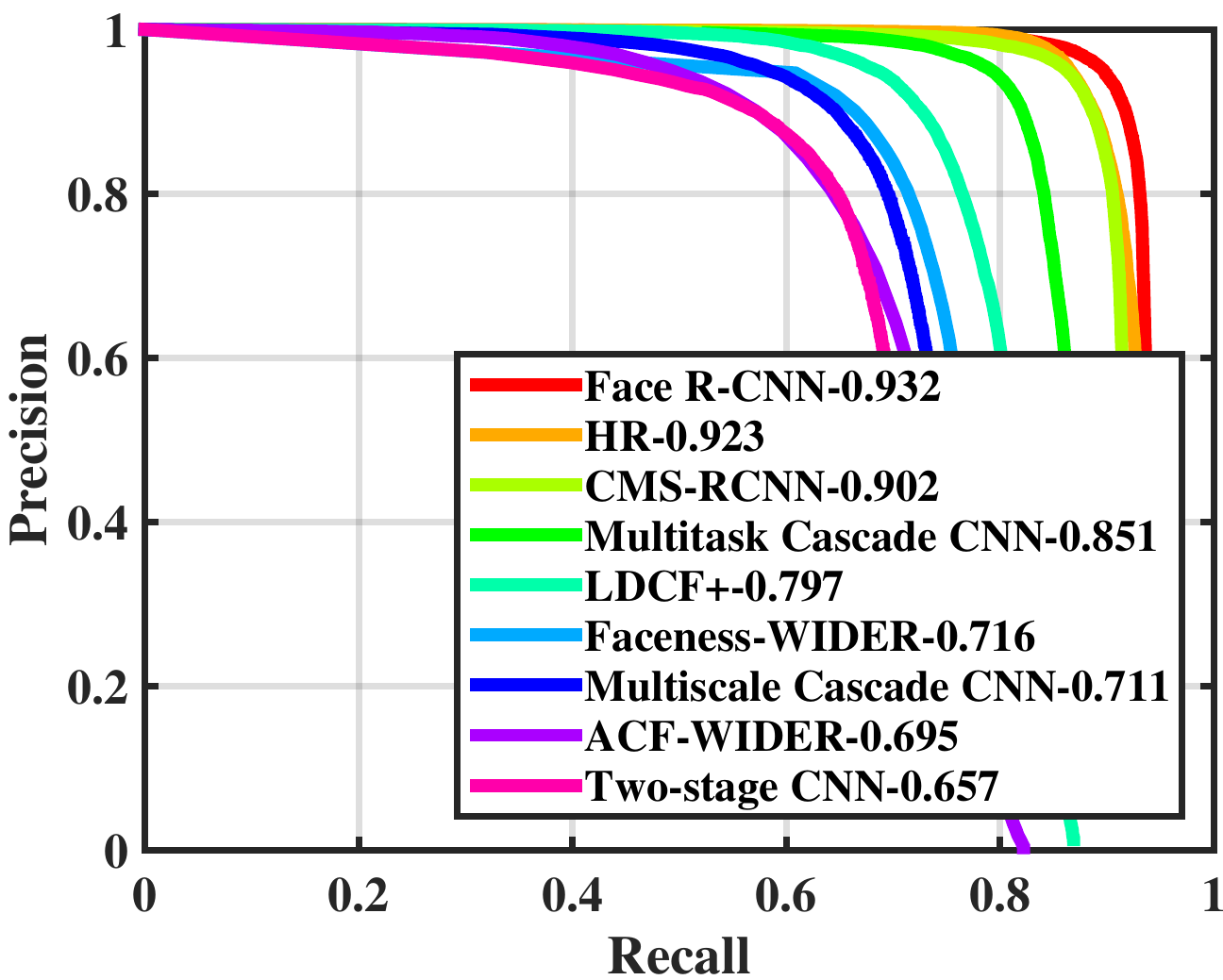}
  }
  \subfigure[Test: medium]
  {
  \label{fig:2:a}
  \includegraphics[width=4cm, keepaspectratio]{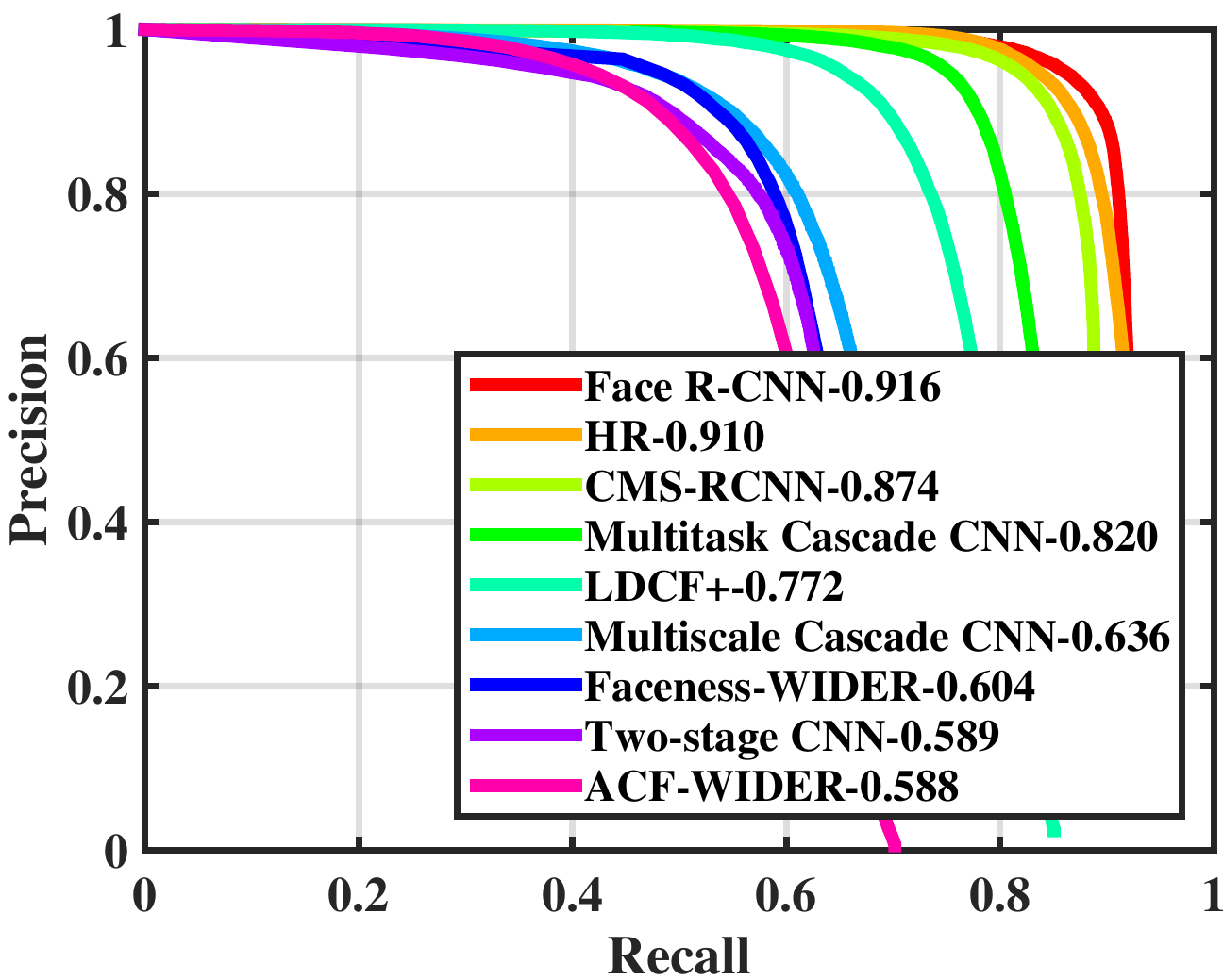}
  }
  \subfigure[Test: hard]
  {
  \label{fig:2:a}
  \includegraphics[width=4cm, keepaspectratio]{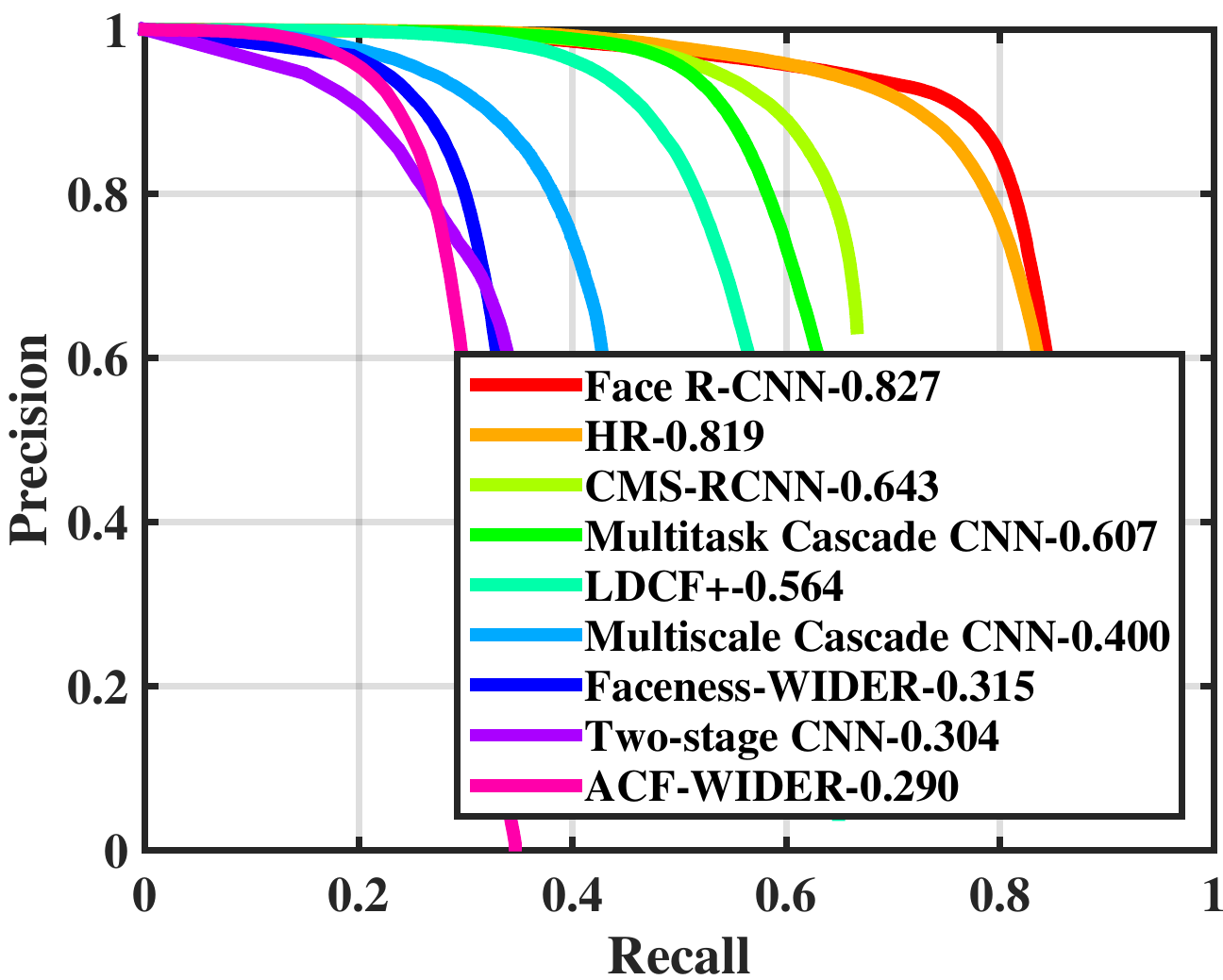}
  }

  \label{fig:2}
  \caption{Precision-Recall curves on WIDER FACE's validation set and test set. All of these methods use the same Scenario-Int criterion \cite{wider}. Our method (Face R-CNN) consistently obtains superior performance over the prior methods across the three subsets (easy, medium, and hard) in the both validation and test sets, thus setting a new state-of-the-art on this challenging benchmark.
}

\end{figure*}

\subsubsection{Multi-Task Loss}

For the RPN stage, we minimize a loss function following the multi-task loss in \cite{FasterRCNN}.
The loss function is the combination of the box-classification loss and the box-regression loss.
The classification loss is softmax loss over background/foreground classes.
The regression loss adopted in our method is SmoothL1, as defined in \cite{FastRCNN}.

For the Fast R-CNN stage, we design a multi-task loss based on the previously introduced center loss. We use the joint supervision of the center loss and the softmax loss
for the classification task, and SmoothL1 Loss for the bounding box regression task. The entire loss function is formulated as:

\begin{equation}\label{1}
 L(p, t, x) = L_{cls}(p, p^*) + {\lambda}L_{reg}(t, t^*) + {\mu}L_c(x).
\end{equation}

\noindent
Here, $p$ is the predicted probability of the proposal region being a face. The ground-truth label $p^*$ is 1 if the proposal
is positive, and is 0 if the proposal is negative. $t$ is a
vector representing the 4 parameterized coordinates
of the predicted bounding box and $t^*$ is that of the
ground-truth. The SmoothL1 Loss is used for regression.
The hyper-parameter $\lambda$, $\mu$ control the balancing weights among
the three terms of the loss.
\vspace{1em}

\subsection{Online Hard Example Mining}

Online Hard Example Mining (OHEM) is a simple yet effective technique for bootstrapping. The key idea is to
collect hard examples that are failed to predict correctly and feed them to
the networks again in order to strengthen the discriminative power.
As the loss can represent how well the current network performs, we sort the generated proposals by their losses and take the top \emph{N}
worst performing examples as the hard examples.

The standard OHEM may encounter the problem of data imbalance, because the selected hard samples may contain an overwhelming number of negative
samples over positive samples (or vice versa). We note that, when using the center loss, it is very crucial to keep the balance of the positive and negative training samples in the learning process. To this end, we apply OHEM on positive samples and negative samples separately and set the ratio of the positive hard samples to the negative hard samples to 1:1 in each mini-batch.

During training, OHEM proceeds with online SGD in an alternating manner. For an SGD iteration,
OHEM is done by performing a forward pass through the current network. Then the selected hard examples are
fed to the network in the next iteration.

\subsection{Multi-Scale Training}

Instead of using a fixed scale for all the training images in the typical Faster R-CNN framework, we design a multi-scale representation for each image by resizing the original image to different sizes during the training process. In this way, the learned model is more adapted to low-resolution faces. In the testing process, multi-scale testing is performed accordingly, and the predicted bounding boxes at different image scales are combined into the final output.

\section{Experiments}

We conduct extensive experiments on two public-domain face detection benchmarks: the WIDER FACE dataset \cite{wider} and the FDDB dataset \cite{fddb}.
The WIDER FACE dataset has a total collection of 393,703 labeled face in 32,203 images,
of which 50\% are used for testing, 40\% for training and 10\% for validation. It is the most challenging face detection benchmark in the public domain.
The FDDB dataset is another well-known face detection benchmark. It contains 5,171 labeled faces in 2,845 images. Some examples of WIDER FACE and FDDB are illustrated in Figure 5.

\subsection{Implementation Details}

We train the detector using VGG19 with the ImageNet pre-trained model.
For the RPN stage, the anchors span a range of multiple scales and aspect ratios.
The anchors with the highest IoU ratios or
having IoU ratios above 0.7 with any ground-truth boxes are assigned positive. The anchors
are assigned negative if their IoU ratios are lower than 0.3.
For the Fast R-CNN stage, we define positive samples having IoU ratios above 0.5
and negative samples having IoU ratios between 0.1 and 0.5, with any ground-truth boxes.
We use the softmax loss and the center loss to jointly guide the learning, as defined in Equation (2).

We adopt non-maximum suppression (NMS) on the RPN proposals.
The IoU of NMS is 0.7 for RPN. After that, a total set of 2000 proposals are produced,
which are then processed with the OHEM method to select
hard examples for Fast R-CNN training.
The mini-batch size is 256 for RPN and 128 for Fast R-CNN, respectively.
Approximate joint training strategy is adopted in our work.

We apply multi-scale training, and the input image is resized with several different
scales. In the testing stage, multi-scale testing is performed as well and the same image scales (as the training stage) are used accordingly.
NMS is applied on the union of predicted boxes using an IoU
threshold of 0.3.

\subsection{Comparison on Benchmarks}

\subsubsection{WIDER FACE}

WIDER FACE has three subsets (Easy, Medium, and Hard) for evaluation
based on different levels of difficulties, as defined in \cite{wider}.
Following the Scenario-Int criterion \cite{wider}, we train our model on the training set of WIDER FACE and conduct
evaluation experiments on the validation set and test set of WIDER FACE.
As shown in Figure 3,
the proposed Face R-CNN consistently wins the 1st place across the three subsets in both the validation set and test set of WIDER FACE,
significantly outperforming the existing results \cite{HR,cmsrcnn,spl,vj2,faceness,wider} on WIDER FACE.

\subsubsection{FDDB}

FDDB is a well-known benchmark released for face detection \cite{fddb}, of which
two evaluation protocol are specified, for 10-fold cross-validation
and unrestricted training respectively. Our experiments strictly follow the protocol for unrestricted training (using the data outside FDDB for training).

\begin{figure}
  \centering

  \subfigure[]
  {
    \label{fig:3:a}
    \includegraphics[width=12cm, keepaspectratio]{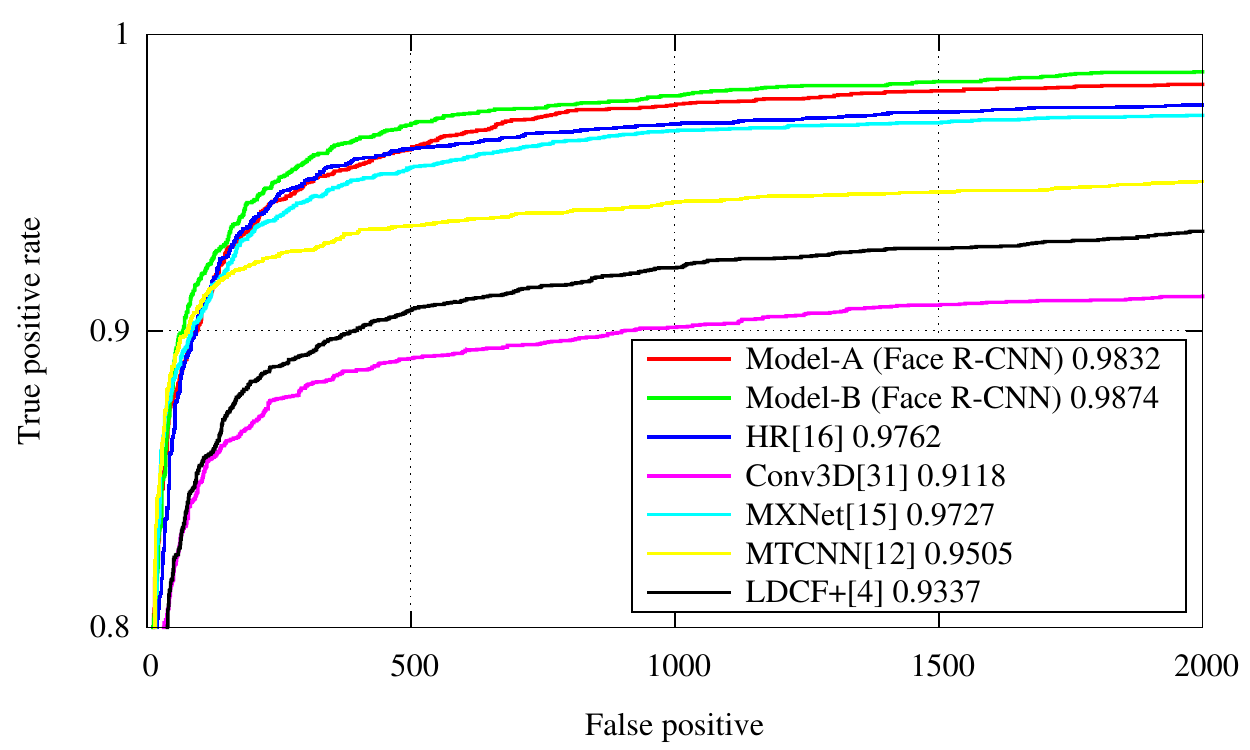}
  }
  \subfigure[]
  {
    \label{fig:3:a}
    \includegraphics[width=12cm, keepaspectratio]{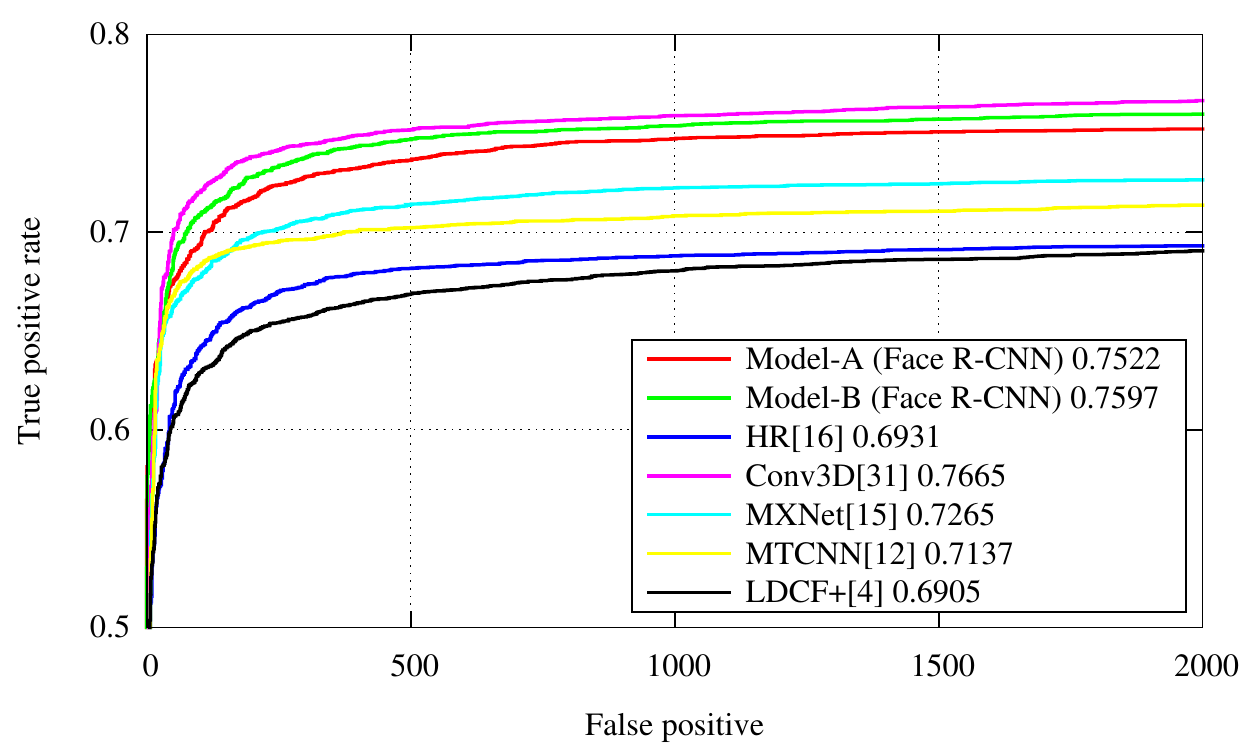}
  }

  \caption{The ROC curves of our method (Face R-CNN) and compared published methods on FDDB. (a) Discrete ROC curve, (b) Continuous ROC curve.
  Model-A is trained using WIDER FACE's training set.
  Model-B is trained using a larger training set by augmenting a private dataset. We also show the true positive rate at 2000 false positives for each method.
  }\label{1}
\end{figure}

 Following \cite{HR}, we use the WIDER FACE's training set to train our model (denoted as Model-A in Figure 4), and compare against the recently published top methods \cite{HR,spl,ldcf,conv3d,mxnet} on FDDB. All of these methods use the protocol for unrestricted training (using the data outside FDDB for training) defined in \cite{fddb}. We plot the discrete ROC curves and continuous ROC curves of these methods in Figure 4, from which we can see that our approach consistently achieves the state-of-the-art performance in terms of both the discrete ROC curve and continuous ROC curve. Especially, our discrete ROC curve is superior to (higher than) the prior best-performing one \cite{HR} by a clear margin. Though our continuous ROC curve is lower than the best-performing one \cite{conv3d}, our discrete ROC curve is much higher than \cite{conv3d}.

Moreover, we enlarge the training set (by augmenting a
privately collected dataset for face detection), and train our face detector (denoted as Model-B). Both the discrete and continuous ROC curves of Model-B are also plotted in Figure 4.
Not surprisingly, the performance of our proposed Face R-CNN can be further improved. Especially, we obtain a true positive rate of 98.74\% of the discrete ROC curve at 2000 false positives, which is a new state-of-the-art among all the published methods on FDDB.

\begin{figure*}
  \centering

  \subfigure[]
  {
    \label{fig:5:a}
    \includegraphics[width=12cm, keepaspectratio]{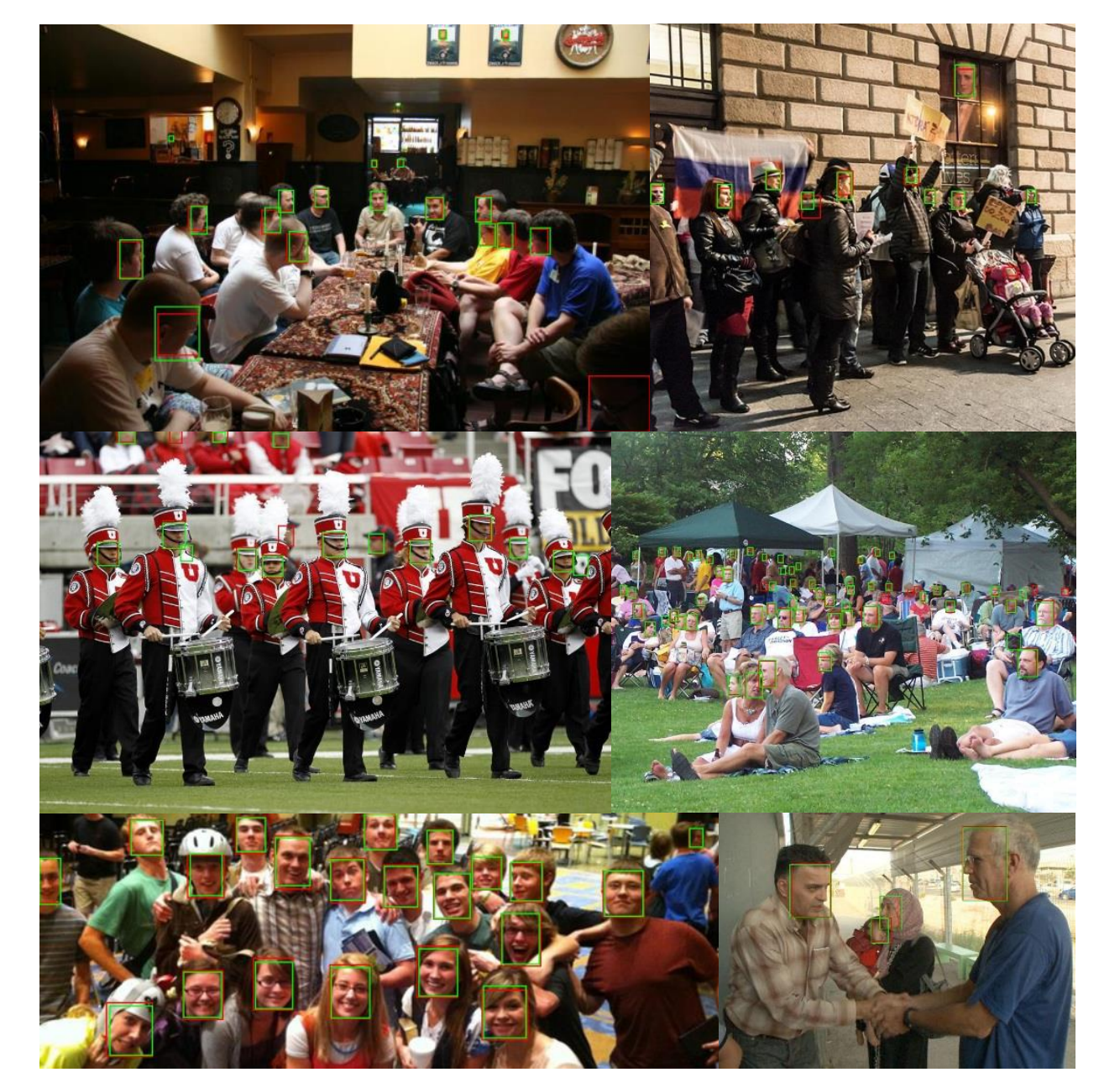}
  }
  \subfigure[]
  {
    \label{fig:5:a}
    \includegraphics[width=12cm, keepaspectratio]{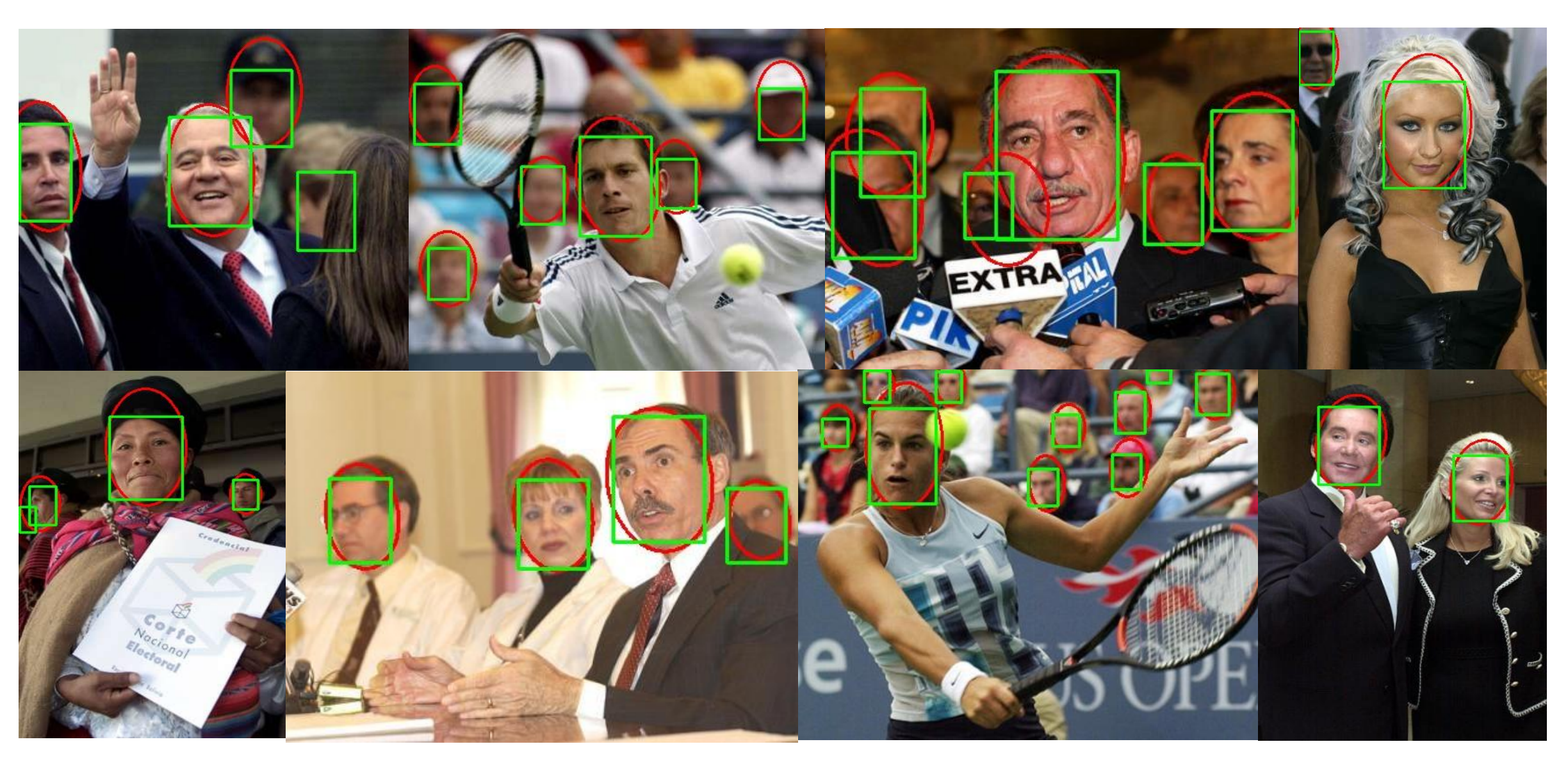}
  }

  \caption{
    Some examples of face detection results. The red boxes or ellipses are the ground-truth annotations.
    The green boxes are the results of our detector. (a) WIDER FACE examples, (b) FDDB examples.
    }\label{1}
\end{figure*}

\section{Conclusion}

In this technical report, we proposed a powerful face detection model called \emph{Face R-CNN} by integrating several newly developed techniques including new multi-task loss function design, online hard example mining, and multi-scale training. The proposed approach is evaluated on the challenging WIDER FACE dataset and FDDB dataset. The experimental results demonstrate the superiority of our approach over the state-of-the-arts.

{\small
\bibliographystyle{plain}
\bibliography{refbib}
}

\end{document}